\documentclass{article}
\usepackage{ijcai26}

\usepackage{times}
\usepackage{soul}
\usepackage{url}
\usepackage[hidelinks]{hyperref}
\usepackage[utf8]{inputenc}
\usepackage[small]{caption}
\usepackage{graphicx}
\usepackage{amsmath}
\usepackage{amsthm}
\usepackage{booktabs}
\usepackage{algorithm}
\usepackage{algorithmic}
\usepackage[switch]{lineno}

\usepackage{amsfonts}
\usepackage{subcaption}
\usepackage{placeins} 
\usepackage{multirow}
\usepackage{threeparttable}
\usepackage{xcolor}
\usepackage{makecell}

\title{Spike-NVPT: Learning Robust Visual Prompts via Bio-Inspired Temporal Filtering and Discretization}

\author{
Qiugang Zhan$^{1,2}$
\and
Anning Jiang$^3$\and
Ran Tao$^{1,2}$\and
Ao Ma$^{1,2,6}$\and
Xiangyu Zhang$^5$\and
Xiurui Xie$^4$\and
Guisong Liu$^{1,2,6,*}$\\
\affiliations
$^1$Complex Laboratory of New Finance and Economics, Southwestern University of Finance and Economics\\
$^2$Engineering Research Center of Intelligent Finance, Ministry of Education\\
$^3$School of Computer Science and Engineering, 
$^4$Laboratory of Intelligent Collaborative Computing, University of Electronic Science and Technology of China\\
$^5$China Mobile Qilu Innovation Research Institute\\
$^6$Kash Institute of Electronics and Information Industry\\
}

\begin{document}

\maketitle

\begin{abstract}
Pre-trained vision models have found widespread application across diverse domains. 
Prompt tuning-based methods have emerged as a parameter-efficient paradigm for adapting pre-trained vision models.
While effective on standard benchmarks, the continuous and dense nature of learned prompts can lead to sensitivity against input noise, as the high-capacity prompts tend to overfit task-irrelevant details.
To address this trade-off, we propose Spike-NVPT, a \textbf{\underline{n}}oise-robust \textbf{\underline{v}}isual \textbf{\underline{p}}rompt \textbf{\underline{t}}uning method.
Specifically, we design a Signal Filtering Layer based on spiking neurons, which uses the integrate-and-fire (IF) mechanism to accumulate task-relevant signals over time and filter transient noise fluctuations.
A subsequent Spike Discretization Unit converts filtered signals into sparse binary prompts.
This discretization acts as a strong regularizer, forcing the model to anchor decision boundaries on the most discriminative and robust features.
Notably, the resulting binary prompts remain static during deployment, ensuring zero additional computational overhead during inference.
Experimental results demonstrate that Spike-NVPT achieves superior robustness performance, with a maximum improvement of 11.2\% over conventional methods, and retains competitive accuracy on clean datasets. 
To the best of our knowledge, this is the first attempt to leverage spiking neurons for fine-tuning traditional artificial neural network (ANN)-based visual models.
\end{abstract}

\section{Introduction}



In recent years, the development of pre-trained vision models has led to remarkable breakthroughs across diverse domains, such as image classification \cite{bazi2021vision,liu2023icmh}, object detection \cite{li2024learning,zheng2021visual}, and semantic segmentation \cite{zhang2024segvit,zhou2024bsbp}.
Leveraging their strong representation capabilities derived from large-scale datasets, these models have been deployed in practical application tasks like autonomous driving \cite{tian2024drivevlm} and intelligent security \cite{sana2024securing}. 
However, in real-world deployments, these models often suffer from various noise interference, such as sensor degradation, adverse weather, or transmission errors \cite{schiappa2023large}.
The noise interference significantly undermines model performance \cite{guo2023robustifying}.
Consequently, enhancing the robustness of these models against natural perturbations is a critical requirement for their practical application.

\begin{figure}[t]
    \centering
    \includegraphics[width=0.47\textwidth]{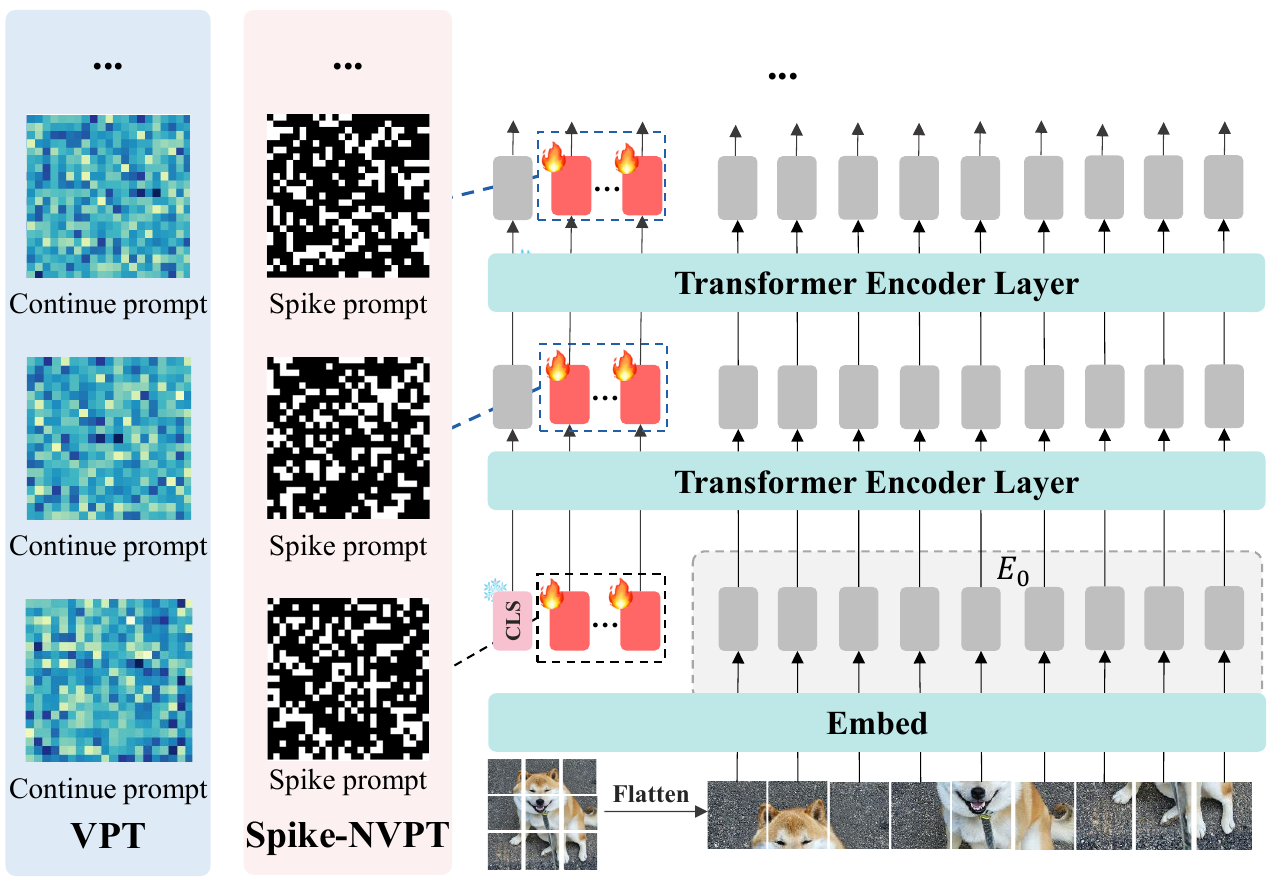}  
    \caption{Schematic diagram of the difference between Spike-NVPT and traditional VPT methods. 
    The prompt generated by Spike-NVPT is in the form of binary spikes.}
    \label{fig: VPT and Spike-NVPT}
\end{figure}

Pre-trained vision models are designed to provide stronger foundational visual capabilities, yet retraining them from scratch to enhance noise robustness for specific tasks is highly inefficient. 
To adapt these massive models efficiently, Parameter-Efficient Fine-Tuning (PEFT) paradigms, such as Visual Prompt Tuning (VPT) \cite{jia2022visual} and ViT-Adapter \cite{chen2023vision}, have been widely adopted.
By injecting a small set of learnable continuous tokens into the frozen backbone, VPT achieves performance comparable to full fine-tuning with a fraction of the parameters.
While these methods successfully reduce training costs, we observe a potential issue rooted in their fundamental design: they typically utilize continuous and dense prompt embeddings. 
This high-capacity parameter space inherently possesses a high degree of freedom that may inadvertently allow the model to overfit to high-frequency noise patterns or task-irrelevant details during fine-tuning. 
Constructing robust fine-tuning mechanisms that can withstand natural noisy interference while maintaining efficiency remains a challenge.

To address this robustness challenge, we look to biological visual systems for inspiration and spiking neural networks (SNNs) show promising potential \cite{li2020robustness,sharmin2020inherent,li2020robustness}. 
The spiking neurons in SNNs process information using discrete spikes via an Integrate-and-Fire (IF) mechanism.
Specifically, a neuron accumulates signals over time and only fires when the accumulated potential exceeds a threshold.
This mechanism functions as a natural temporal filter to exclude transient, random noise fluctuations while retaining persistent, task-relevant signals \cite{wang2023new,ding2025neuromorphic}.
Driven by these studies, we consider that incorporating these bio-inspired filtering and discretization dynamics into the prompt learning process can mitigate the noise sensitivity associated with continuous prompts.

Building on these analyses, we propose Spike-NVPT, a novel noise-robust visual prompt tuning method inspired by SNNs.
Rather than deploying a full SNN for inference, we leverage SNN dynamics strategically during the training phase to "distill" robust prompts.
Specifically, Spike-NVPT comprises two key components: the \textbf{Signal Filtering Layer} and the \textbf{Spike Discretization Unit}. 
In the tuning process, we first randomly initialize a set of continuous prompts. 
These prompts are then fed into the Signal Filtering Layer, which accumulates and filters prompt signals in the temporal domain.
The filtered information is subsequently passed to the Spike Discretization Unit, which converts it into sparse binary spiking prompts.
Crucially, the iterative SNN simulation is only required during training. 
For inference, the learned binary prompts are fixed as static task-specific inputs, ensuring that our method incurs zero additional computational overhead compared to standard VPT.
Figure \ref{fig: VPT and Spike-NVPT} shows the difference between Spike-NVPT and traditional VPT.

A key point of our approach lies in where the SNN dynamics are applied. 
Typically, the discrete, binary nature of SNNs is considered a drawback for processing input images due to the loss of precise information.
However, Spike-NVPT applies SNN-based binarization exclusively to the prompts, leaving the input image embeddings continuous and dense.
We transform this "information loss" into a beneficial regularization mechanism.
The binary prompts act as strict constraints that filter out noise-sensitive variables, guiding the model to capture the most discriminative features rather than fitting every detail of the input.
The main contributions of this paper can be summarized as follows:
\begin{itemize}
    \item For the first time, this work explores the feasibility of using SNN to fine-tune ANN-based pre-trained vision models.
    \item We propose the Spike-NVPT method, which incorporates a Signal Filtering Layer and a Spike Discretization Unit, thereby enhancing the robustness of the fine-tuned model to noisy datasets.
    \item Our method outperforms the different types of fine-tuning approaches like VPT, LoRA, and Dynamic Tuning on noisy datasets. The spike prompt has been demonstrated to consistently guide models in extracting robust features by layer.
\end{itemize}

\section{Related Work and Preliminary}


\subsection{Embedding-based Fine-tuning in Computer Vision}
With the advancement of pre-trained vision models, efficient fine-tuning for downstream tasks has become a key research focus. 
Visual prompt tuning paradigm integrates a set of learnable parameters, termed prompts, into the embedding space of each layer in the pre-trained model.
Such embedding-level fine-tuning can effectively reduce training overhead on specific downstream tasks.

VPT \cite{jia2022visual} is the first attempt at embedding-level fine-tuning in pre-trained visual models and shows remarkable performance. 
Then, Pro-Tuning \cite{nie2023pro} designs lightweight prompts, including three convolutional layers, which generate specific discriminative prompts.
CVP \cite{tsai2023convolutional} is proposed to achieve the perception of out-of-distribution samples without labeling.
IDPT \cite{zha2023instance} presents a new instance-aware dynamic prompt-tuning strategy for pre-trained point cloud models.
A two-stage prompt learning approach is designed in LPT \cite{dong2023lpt} to reduce the model's overfitting to certain features of long-tailed data.
ViPT \cite{zhu2023visual} develops a visual prompt-tuning method that learns modality-relevant prompts to adapt pre-trained base models to various downstream multimodal tracking tasks.
Recently, LION \cite{wang2024lion} has extended the idea by inserting two implicit layers before the input and the output of the model backbone. 

While embedding-level fine-tuning effectively reduces training overhead, most existing methods operate within a continuous and dense parameter space. 
Such high-capacity prompts lack intrinsic structural constraints, making them prone to overfitting high-frequency noise patterns or task-irrelevant details. 
Approaches for fine-tuning prompts to enhance robustness against natural noise still require further exploration.

\subsection{Noise Robustness of Spiking Neural Networks}
Spiking neural networks are a novel type of neural network inspired by biological neural systems. 
In contrast to traditional ANNs, which process information using continuous activation values, SNNs utilize discrete spike signals to process information in the temporal domain \cite{maass1997networks}.

The basic units of SNNs are spiking neurons, of which the integrate-and-fire (IF) model is one of the most famous. 
Specifically, the membrane potential $V(t)$ of an IF neuron accumulates the input current over time, and when it exceeds the threshold $V_{th}$, the neuron emits a spike \cite{gerstner2002spiking}.
The membrane potential $V(t)$ at the time step $t$ is derived formally as follows,
\begin{equation}
    V(t) = V'(t-1) + I(t),
\label{eq: IF}
\end{equation}
where $I(t)$ represents the input current fed into the IF neuron at the time step $t$; $V'(t-1)$ is the subthreshold membrane potential at time $t-1$. 
Once $V(t)$ exceeds the threshold $V_{th}$, the neuron fires a spike, defined as, 
\begin{equation}
\begin{aligned}
    S(t) &= sign\left(V(t) - V_{th}\right) \\
    &=
    \begin{cases} 
        1, & V(t) - V_{th} \geq 0, \\
        0, & V(t) - V_{th} < 0,
    \end{cases}
\end{aligned}
\label{eq: firing}
\end{equation}
where $S(t)$ represents the output of neuron.
Then, the subthreshold membrane potential is updated according to the firing state as follows, 
\begin{equation}
    V'(t) = V(t) \cdot \left(1 - S(t)\right) + V_{reset} \cdot S(t),
\label{eq: reset}
\end{equation}
where $V_{reset}$ means the predefined resting potential.

\begin{figure*}[t]
    \centering
    \includegraphics[width=0.9\textwidth]{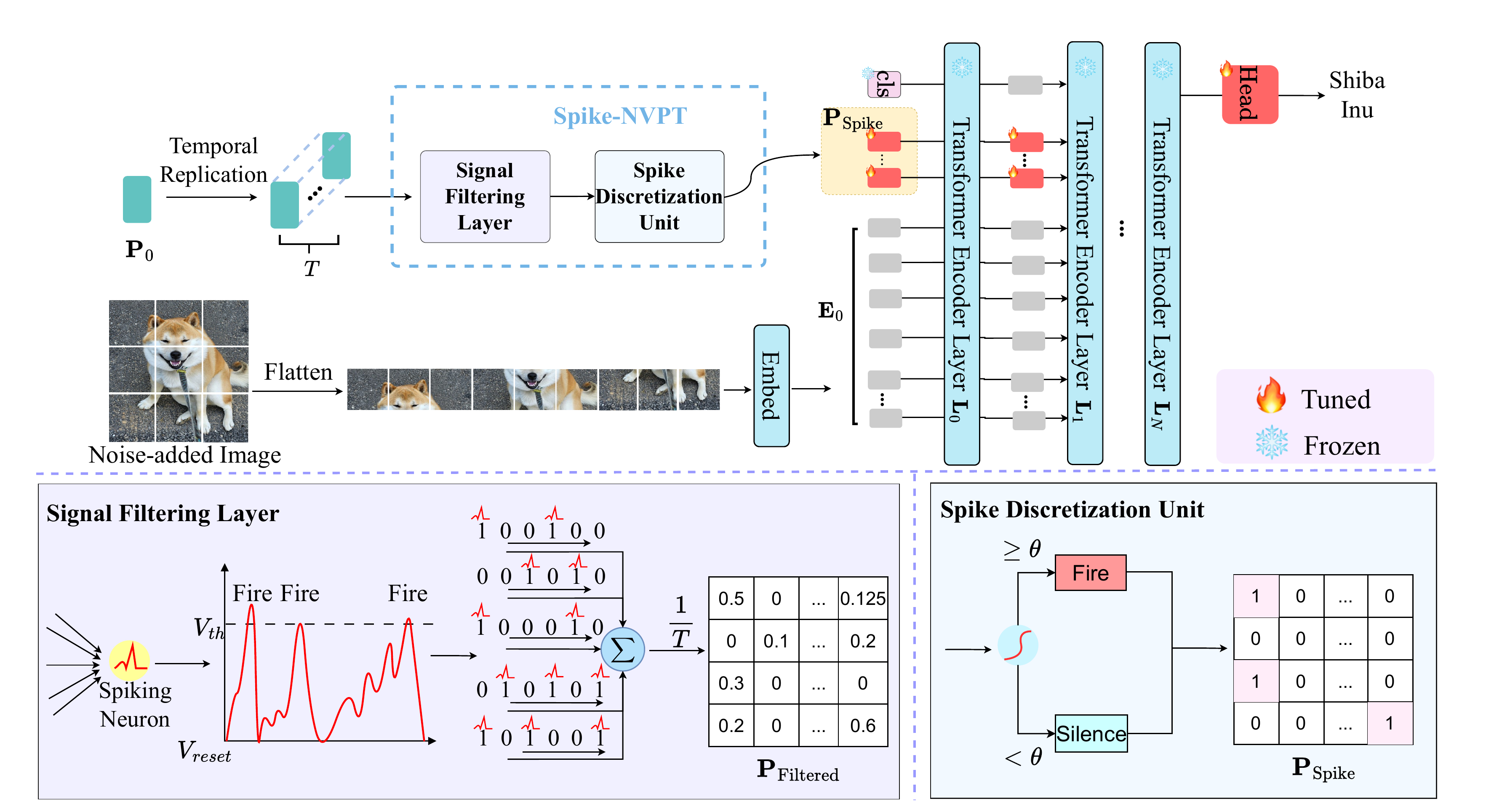}  
    \caption{The overall architecture of Spike-NVPT. 
    It consists of two main components: 
    (1) Signal Filtering Layer. Spiking neurons in this layer filter the important information from the initial continuous prompt; 
    and (2) Spike Discretization Unit. The extracted information is further translated into spike form to improve noise resistance, and sparse features are utilized to provide regular constraints on the classification header.}
    \label{fig: Spike-NVPT}
    \vspace{1em}
\end{figure*}

The spiking mechanism and its reliance on temporal dynamics enhance SNN robustness in noisy environments \cite{ding2024robust}, with discrete encoding and nonlinear spiking neuron activation being key contributors \cite{sharmin2020inherent}. 
Recent advances have seen spiking neuron-based models \cite{zhan2024two,zhan2024spiking} widely applied to improve noise robustness \cite{kundu2021hire,ding2022snn,ding2025neuromorphic}.

Building on these insights, we propose to leverage the mechanism of SNNs for robust fine-tuning.
we take the SNN dynamics as a regularization strategy during training rather than a deployment architecture.
The sparse binary prompts are trained as task-specific priors without losing the original information of the input images.

\section{Method}

\begin{figure}[t]
    \centering
    \includegraphics[width=0.45\textwidth]{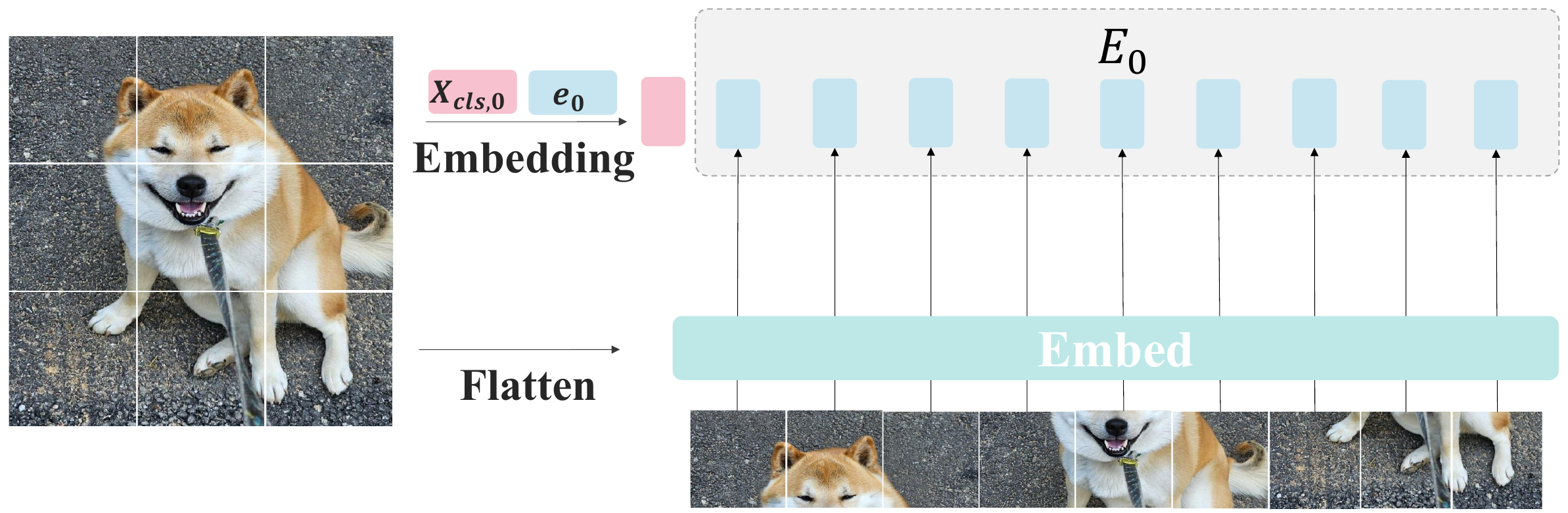}  
    \caption{Illustration of generating patch embeddings from the input image. The image is divided into patches, and corresponding embeddings are generated for each patch, which are then used in subsequent processing steps.}
    \label{fig: Patch Embedding}
\end{figure}

\subsection{Overall Architecture} \label{sec: Overall Architecture}
The overall architecture is illustrated in Figure \ref{fig: Spike-NVPT}.
The input image is first divided into $m$ fixed-size image patches, which are clean in the fine-tuning process and are noisy in the test phase. 
These patch are embedded into corresponding vectors denoted by $\mathbf{E}_0=\left\{\mathbf{e}_0^j\right\}_{j=1}^{m}$, where $\mathbf{e}_0^j$ represents the embedding of the $j$-th patch and its dim is $D$, i.e., $\mathbf{e}_0^j \in \mathbb{R}^D$. 
In addition, a learnable classification [CLS] token denoted by $\mathbf{x}_{i, cls} \in \mathbb{R}^D$ is added before these embeddings, as shown in Figure \ref{fig: Patch Embedding}. 
Given a pre-trained vision model with $N$ transformer encoder layers, the whole model can be formulated as, 
\begin{equation}
\begin{aligned}
    \relax [\mathbf{x}_{i+1,cls}, \mathbf{E}_{i+1}] = L_{i+1, T}\left([\mathbf{x}_{i, cls}, \mathbf{E}_{i}]\right), \\
    i = 1, 2, \dots, N,
\end{aligned}
\label{eq: vit}
\end{equation}
\begin{equation}
    \mathbf{y} = Head(\mathbf{x}_{N, cls}),
\end{equation}
where $[\mathbf{x}_{i, cls}, \mathbf{E}_{i}]$ present the input of the $i+1$-th layer, including the [CLS] token embedding and patch embeddings in the $i$-th layer, respectively. $L_{i, T}$ is the $i$-th transformer layer and $Head(\cdot)$ means the classification head.

In the fine-tuning phase, we follow the classic visual prompt tuning paradigm.
Firstly, we define $N+1$ learnable prompts $\left\{\mathbf{P}_{ i}\right\}_{i=0}^{N}$ for each layer, including the input embedding layer.
Each prompt $\mathbf{P}_{i} = \left\{\mathbf{p}_i^k \in \mathbb{R}^D \right\}_{k=1}^{K}$ are initialized randomly, where $K$ is the token number of prompt.
For the $i$-th layer, we feed the prompt $\mathbf{P}_i$ into the Signal Filtering Layer $L_{F}(\cdot)$. 
The Signal Filtering Layer is implemented by SNN, which can preliminarily filter information on the randomly initialized prompt $\mathbf{P}_i$ and output the filtered prompt $\mathbf{P}_{i, \text{Filtered}}$.
Then, $\mathbf{P}_{i, \text{Filtered}}$ is fed into the Spike Discretization Unit $U_{S}(\cdot)$ to generate the binary spiking prompts $\mathbf{P}_{i, \text{Spike}}$. 
These spiking prompts will be concatenated with the input embedding as the cues for pre-trained models.
Formally, our whole architecture is described as, 
\begin{equation}
\begin{aligned}
    \relax [\mathbf{x}_{i+1,cls}, \_, \mathbf{E}_{i+1}] = L_{i+1, T}([\mathbf{x}_{i, cls}, \mathbf{P}_{i, \text{Spike}}, \mathbf{E}_{i}]), \\
    i = 0, 1, \dots, N,
\end{aligned}
\end{equation}
\begin{equation}
     \mathbf{P}_{i, \text{Filtered}} = L_{F}\left(\mathbf{P}_i\right),
\end{equation}
\begin{equation}
     \mathbf{P}_{i, \text{Spike}} = U_{S}\left(\mathbf{P}_{i, \text{Filtered}}\right),
\end{equation}
where the symbol $\_$ corresponds to the output of $\mathbf{P}_{i, \text{Spike}}$ and will not be used subsequently.




During training, only the task-specific parameters within the prompt embedding set are fine-tuned, i.e., $\mathbf{P}_i$, while the backbone network parameters are frozen.

\subsection{Signal Filtering Layer}\label{sec: Signal Filtering Layer}
The Signal Filtering Layer of Spike-NVPT processes input signals through spiking neurons, which accumulate membrane potential over time and emit spikes upon surpassing a defined threshold. 
Utilizing rate coding, this layer computes the average spike firing rate of spiking neurons over a defined time window to extract and compress signal features. 
This design allows the layer to suppress noise in certain situations while generating high-quality temporal features that are task-relevant. 
The optimized signal representation is then provided as input for the subsequent unit.

The Signal Filtering Layer $L_{F}(\cdot)$ consists of $K \times D$ IF nodes, the same as the number of elements in the prompt $\mathbf{P}_i$.
$L_{i, F, k, d}(\cdot)$ denotes the IF node corresponding to the $d$-th element of the $k$-th token of the prompt $\mathbf{P}_i$.
$L_{i, F, k, d}(\cdot)$ receives the randomly initialized prompt $\mathbf{p}_i^k[d]$ as the input current described in Eq. \eqref{eq: IF}.
At each time step $t \in \{1,2, \cdots, T\}$, the membrane potential of the $L_{i, F, k, d}(\cdot)$ is, 
\begin{equation}
    V_{i, F, k, d}(t) = V'_{i, F, k, d}(t-1) + \mathbf{p}_i^k[d].
\end{equation}
When the accumulated membrane potential exceeds the threshold $V_{th}$, the neuron emits a spike; otherwise, the neuron remains silent, following Eq. \eqref{eq: firing}.
The spike firing state of the IF node $L_{i, F, k, d}(\cdot)$ at time step $t$ is denoted by $S_{i, F, k, d}(t) \in [0, 1]$.
Subsequently, the subthreshold membrane potential $V'_{i, F, k, d}(t)$ is calculated by Eq. \eqref{eq: reset}.

By accumulating spikes over multiple time steps, the model integrates persistent task-related signals in the temporal dimension, thus enhancing the representation of task-specific features. 
Therefore, the output spikes $S_{i, F, k, d}(t)$ are accumulated over $T$ time steps and normalized to generate a filtered prompt $\mathbf{p}_{i, \text{Filtered}}^k[d]$, which represents the signal intensity.
This process is formulated by, 
\begin{equation}
    \mathbf{p}_{i, \text{Filtered}}^k[d] = \frac{1}{T}\sum_{t=1}^TS_{i, F, k, d}(t).
\end{equation}
Therefore, through the Signal Filtering Layer, the information of the initial prompt $\mathbf{P}_i$ is filtered to get $\mathbf{P}_{i, \text{Filtered}} \in \left\{\frac{1}{T}, \frac{2}{T}, \dots, \frac{T}{T}\right\}^{K\times D}$.

\subsection{Spike Discretization Unit}\label{sec: Spike Discretization Unit}
The Spike Discretization Unit serves as a quantization bottleneck, transforming the filtered continuous signals $\mathbf{P}_{i, \text{Filtered}}$ into a sparse binary code $\mathbf{P}_{i, \text{Spike}}$.
This design departs from the continuous representations used in traditional prompt tuning by imposing a sparse binary prior. 
We take the ``firing'' activation as a strong regularizer. 
By forcing the prompt values to collapse to extreme points (0 or 1), we can prevent the model from fine-tuning its decision boundaries based on noise-prone variations in the prompt space. 
Instead, it compels the backbone to focus only on the most significant features that have survived the temporal filter.


Based on the above analysis, the Spike Discretization Unit is designed to translate the $\mathbf{P}_{i, \text{Filtered}}$ into a binary spiking prompt $\mathbf{P}_{i, \text{Spike}}$, which aims to translate the information obtained from the Signal Filtering Layer into a more robust spike form.


The Spike Discretization Unit amplifies relevant signals to form full spikes while suppressing irrelevant signals to a silent state, thereby making the embeddings more effective and better integrated within the entire set of input features. Specifically, for the $\mathbf{p}_{\text{Filtered}}^k[d]$ generated by the Signal Filtering Layer, the Spike Discretization Unit amplifies or mutes signals as follows, 
\begin{equation}
\begin{aligned}
    \mathbf{p}_{i, \text{Spike}}^k[d] 
    &= sign\left(\mathbf{p}_{i, \text{Filtered}}^k[d] - \theta\right) \\
    &=
    \begin{cases} 
        1, & \mathbf{p}_{i, \text{Filtered}}^k[d] - \theta \geq 0, \\
        0, & \mathbf{p}_{i, \text{Filtered}}^k[d] - \theta < 0,
    \end{cases}
\end{aligned}
\label{eq: unit}
\end{equation}
where the parameter $\theta$ is a threshold that determines the boundary between the effective and silent signals. 
However, since the indicator function $sign(\cdot)$ is inherently a discontinuous step function, it leads to an unavailable gradient during backpropagation. 
To overcome the limitation, we use the gradient of the arctangent function as a surrogate function, Eq. \eqref{eq: unit}. 
The surrogate function is defined as, 
\begin{equation}
    \frac{\partial \mathbf{p}_{i, \text{Spike}}^k[d]}{\partial \mathbf{p}_{i, \text{Filtered}}^k[d]} = \frac{\alpha}{2 \left(1 + \left( \frac{\pi}{2} \alpha \cdot (\mathbf{p}_{i, \text{Filtered}}^k[d] - \theta))\right) ^2\right)},
\label{eq: surrogate gradient}
\end{equation}
where parameter $\alpha > 0$ is the gradient scaling factor to control the steepness of the surrogate function. 

By employing this strategy, this unit amplifies meaningful signals and minimizes the impact of non-relevant signals. 

\section{Experiments}

\subsection{Experimental Setup}
\textbf{\textit{Datasets.}} 
We evaluate our method on four types of scenes, including natural objects, fine-grained categories, texture images, and synthetic visual environments.
The natural scene includes CIFAR-100 \cite{krizhevsky2009learning} and Caltech-101 \cite{fei2007learning} datasets.
The fine-grained scene consists of Oxford-IIIT Pets (Pets) \cite{parkhi2012cats} and Oxford Flowers 102 (Flowers) \cite{nilsback2008automated} datasets.
The texture scene has DTD \cite{cimpoi2014describing} and RESISC45 \cite{Cheng_2017} datasets.
We use the DMLab \cite{zhai2019large} dataset to represent the synthetic scene.
Dataset descriptions are provided in the supplementary material.

\begin{figure}[t]
    \centering
    \includegraphics[width=0.35\textwidth]{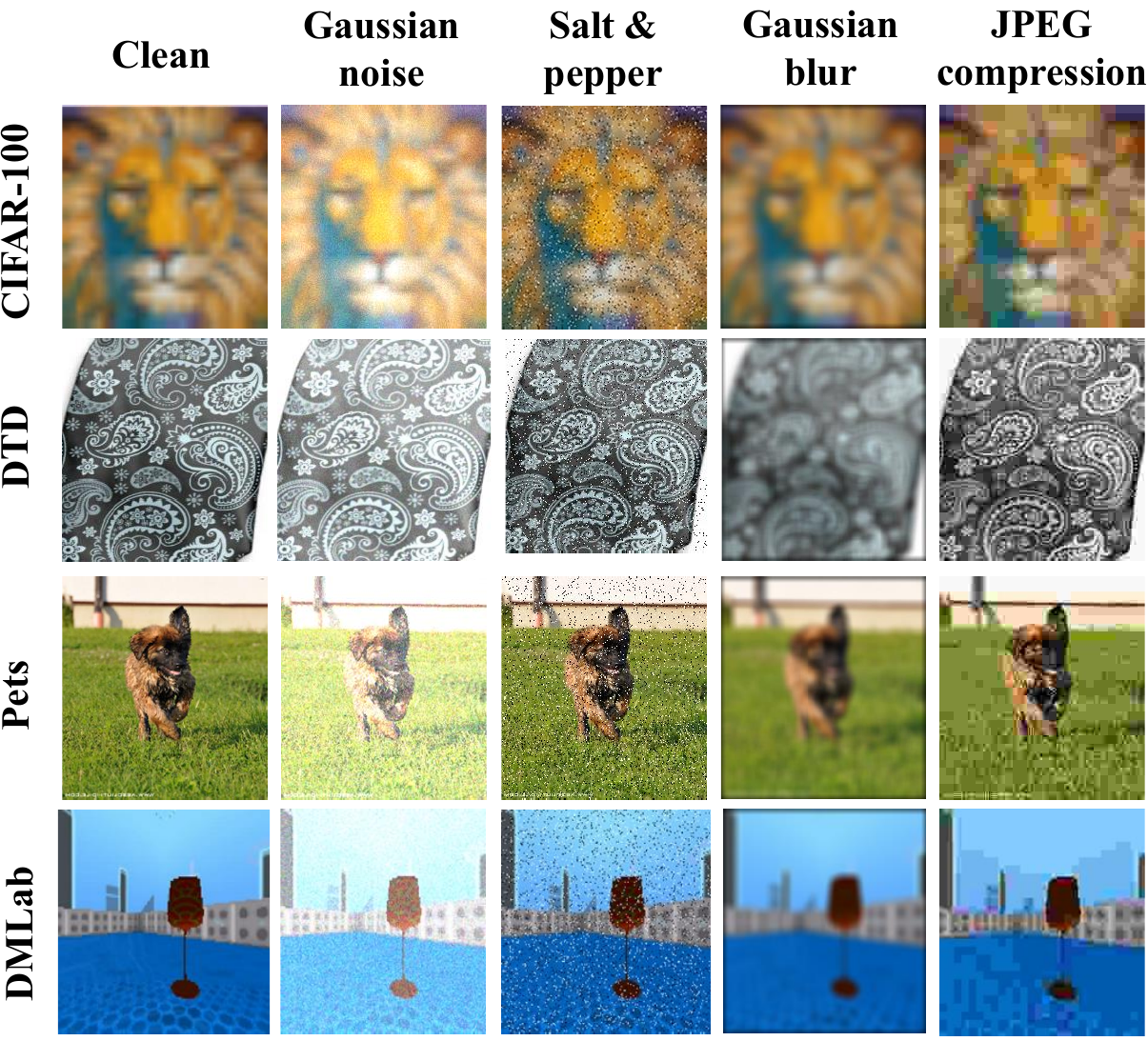}  
    \caption{Images of CIFAR-100, DTD, Pets, and DMLab with varying noise intensities.}
    \label{fig: example}
\end{figure}

\noindent\textbf{\textit{Noise setting.}} 
We introduce two basic High-frequency natural noises, Gaussian and salt-and-pepper, with varying intensities to validate noise robustness. 
We set the mean of Gaussian noise to \{0.1, 0.2, 0.3, 0.4\} with a standard deviation of 0.1.
The salt and pepper rates are set to \{0.01, 0.02, 0.03, 0.04\}.
Additionally, we introduce two complex structural natural perturbations, Gaussian Blur and JPEG compression (JPEG Comp.).
The JPEG compression quality parameter quality is set to \{20, 15, 10, 5\}. 
The smaller the value, the lower the image quality.
The blurring degree parameter sigma range is set to \{2, 3, 4, 5\}, with higher values resulting in greater blurring.
Figure \ref{fig: example} shows example noisy images.

\noindent\textbf{\textit{Baselines.}} 
We take parameter-efficient fine-tuning methods of different types with open-source code for comparison, including adapter, low-rank optimization, and prompt, specifically Dynamic Tuning \cite{zhao2024dynamic}, LoRA \cite{hu2022lora}, and VPT \cite{jia2022visual}.
Additionally, due to the scarcity of noise-robust fine-tuning methods, we introduce the adversarial robust fine-tuning method EdgeNet \cite{li2024harnessing} as a comparison.
It constructs an adapter that utilizes image edge information to generate robust features. 

\noindent\textbf{\textit{Training details.}} 
To ensure the fairness of the experiments, all methods are based on the same pre-trained model, ViT-B/16, as the backbone network. 
The SGD optimizer with a batch size of 64 is employed, and the model is trained for 100 epochs across all datasets. 
In line with VPT \cite{jia2022visual}, we set prompt lengths to 10 for CIFAR-100, Caltech-101, DTD, Flowers, and Pets, while DMLab and RESISC45 adopt 50. 
Initial learning rate and weight decay are set to 1.5 and 0.001, respectively, determined based on the range identified in \cite{jia2022visual}.
Time steps $T$ are uniformly set to 16, reflecting input-sequence complexity and temporal-pattern learning capacity \cite{valadez2017step}.
Following previous research \cite{neftci2019surrogate,liao2023convolutional}, the parameter $\alpha$ in Eq. 13 is set to 2.
Both the firing threshold $V_{th}$ in the Signal Filtering Layer and $\theta$ in the Spike Discretization Unit are set to 0.01.

The hyperparameters of the comparison methods are set using the corresponding settings of the original paper.
The initial version of LoRA did not experiment on visual models; therefore, we used the hyperparameters of Dynamic Tuning for LoRA training \cite{zhao2024dynamic}.

The experiments are conducted on an NVIDIA GeForce RTX 4090 GPU, using the SpikingJelly (version 0.0.0.0.14) \cite{doi:10.1126/sciadv.adi1480} for spiking neural network simulation, with PyTorch 2.2.2 as the deep learning backend.


\begin{table*}[htbp]
    \centering
    \renewcommand\arraystretch{1.0}
    \resizebox{0.9\linewidth}{!}{
        \begin{tabular}{lccccc|ccccc|c}
        \toprule
        \multirow{2}{*}{Methods} & \multicolumn{5}{c|}{Natural (CIFAR-100, Caltech-101)} & \multicolumn{5}{c|}{Fine-grained (Pets, Flowers)} & \multirow{2}{*}{Params (M)} \\
        \cline{2-6} \cline{7-11}
        & Clean & \makecell{Gaussian\\noise} & \makecell{Salt-and-\\pepper} & \makecell{JPEG\\Comp.} & \makecell{Gaussian\\blur} & Clean & \makecell{Gaussian\\noise} & \makecell{Salt-and-\\pepper} & \makecell{JPEG\\Comp.} & \makecell{Gaussian\\blur} & \\
        \midrule
        VPT & \underline{86.15} & 55.40 & 50.35 & 65.25 & 68.60 & 92.85 & 83.40 & 78.20 & 80.15 & 81.15 & 0.17 \\
        LoRA & 81.95 & 59.25 & 54.10 & 66.20 & \underline{72.75} & 93.69 & 84.30 & 79.15 & \underline{81.15} & \underline{82.30} & 0.67 \\
        EdgeNet & 80.50 & \underline{63.20} & 53.70 & 59.50 & 66.60 & 93.61 & 85.55 & 78.00 & 76.20 & 74.80 & 33.51 \\
        Dynamic-tuning & 83.40 & 59.70 & \textbf{60.90} & \underline{68.20} & 71.40 & \textbf{95.45} & \textbf{86.80} & \underline{81.90} & 80.90 & 73.95 & 1.28 \\
        \textbf{Spike-NVPT (ours)} & \textbf{87.35} & \textbf{63.90} & \underline{58.40} & \textbf{69.20} & \textbf{75.00} & \underline{94.58} & \underline{86.70} & \textbf{82.70} & \textbf{82.85} & \textbf{85.60} & \textbf{0.17} \\
        \hline
        & \multicolumn{5}{c|}{Texture (DTD, RESISC45)} & \multicolumn{5}{c|}{Synthetic (DMLab)} & Average \\
        \cline{2-6} \cline{7-12}
        VPT & 77.35 & 54.60 & \underline{50.20} & 59.30 & 47.45 & 43.50 & 39.70 & 39.10 & 39.10 & 33.00 & 61.24 \\
        LoRA & 72.30 & 53.40 & 42.40 & 55.35 & \underline{48.05} & 47.20 & 39.40 & 40.20 & 39.70 & \textbf{38.20} & 61.55 \\
        EdgeNet & 71.10 & 43.80 & 39.70 & 42.90 & 29.45 & 39.90 & 31.20 & 28.00 & 32.90 & 27.40 & 55.90 \\
        Dynamic-tuning & \textbf{80.15} & \underline{57.55} & 50.10 & \underline{59.85} & 50.20 & \textbf{50.00} & \textbf{44.30} & \textbf{44.10} & \textbf{44.30} & \underline{37.40} & \underline{64.03} \\
        \textbf{Spike-NVPT (ours)} & \underline{77.70} & \textbf{59.45} & \textbf{55.00} & \textbf{60.30} & \textbf{59.25} & \underline{47.60} & \underline{40.90} & \underline{40.90} & \underline{40.10} & 35.20 & \textbf{65.13} \\
        \bottomrule
        \end{tabular}
    }
    \caption{Performance comparison across different datasets and noise types. \textbf{Bold font} and \underline{underline} denote the best and the second-best performance, respectively. 
    Subsequent tables use the same markup.
    The average column indicates the mean value across all seven datasets.}
    \label{tab: performance}
\end{table*}

\subsection{Experimental Results}
In this section, we compare our method with other fine-tuning methods across four scenes.
For each dataset, we conduct tests at all noise levels across the four noise types and average the results by noise type.
To facilitate presentation and comparison, we further average the results by test scene, as shown in Table \ref{tab: performance}.
The complete test results are presented in the supplementary material.

\begin{figure}[htbp]
  \centering
  \includegraphics[width=0.9\linewidth]{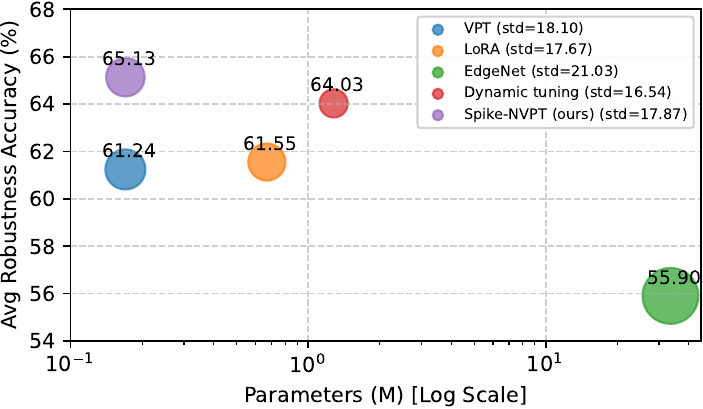}
  \caption{The trade-off result between average performances and parameter scales. The scatter plot size represents the standard deviation of results across both clean and noisy test sets, with specific values documented in the legend.}
  \label{fig: trade-off}
\end{figure}

\textbf{\textit{Average robustness analysis.}}
According to the results, Spike-NVPT demonstrates a significant advantage in average noise robustness compared to other fine-tuning baselines.
Specifically, in terms of average robustness accuracy, our method achieves 65.13\%, outperforming the original VPT 61.24\% by a substantial margin of 3.89\%.
As one of the most commonly used fine-tuning methods for large language models, LoRA demonstrates comparable robustness to VPT.
EdgeNet achieves only an average accuracy of 55.90\%, indicating that adversarial robustness methods are not universally effective against noise.
Dynamic tuning achieved the second-highest performance at 64.03\%, but as an adapter-based approach, it also introduces more training parameters.
Figure \ref{fig: trade-off} illustrates the trade-off between different methods in terms of average performance and parameter scale.
Spike-NVPT achieves the best cost-performance ratio and acceptable robustness fluctuations.

\textbf{\textit{Parameter-efficient analysis.}}
A common concern with robust fine-tuning is the potential trade-off with clean accuracy.
However, our results indicate that Spike-NVPT incurs no performance penalty on clean data, and it achieves the best or second-best accuracy.
This finding supports our claim that the 'information loss' caused by the SNN binarization is a beneficial feature selection rather than a destructive loss. 
By filtering out redundant continuous signals, our model focuses on the most discriminative semantic features, thereby enhancing generalization even in noise-free scenes.

\textbf{\textit{Performance on synthetic scene analysis.}}
In synthetic scenes, i.e., the DMLab dataset, our Spike-NVPT slightly trails behind Dynamic-tuning.
This is because the backbone, ViT, is trained on real-world datasets, and synthetic images exhibit a significant domain shift relative to it.
Dynamic Tuning, with its vast parameter space and dynamic adjustment mechanism, possesses superior domain adaptation capabilities, enabling it to fit such enormous domain differences.
Despite having a larger number of parameters, EdgeNet fails to achieve good performance due to the difference between adversarial robustness and noise robustness.
Spike-NVPT achieves the best results among fine-tuning methods of comparable scale, demonstrating its broad applicability.

\begin{figure*}[htbp]
  \centering
  \includegraphics[width=1.0\linewidth]{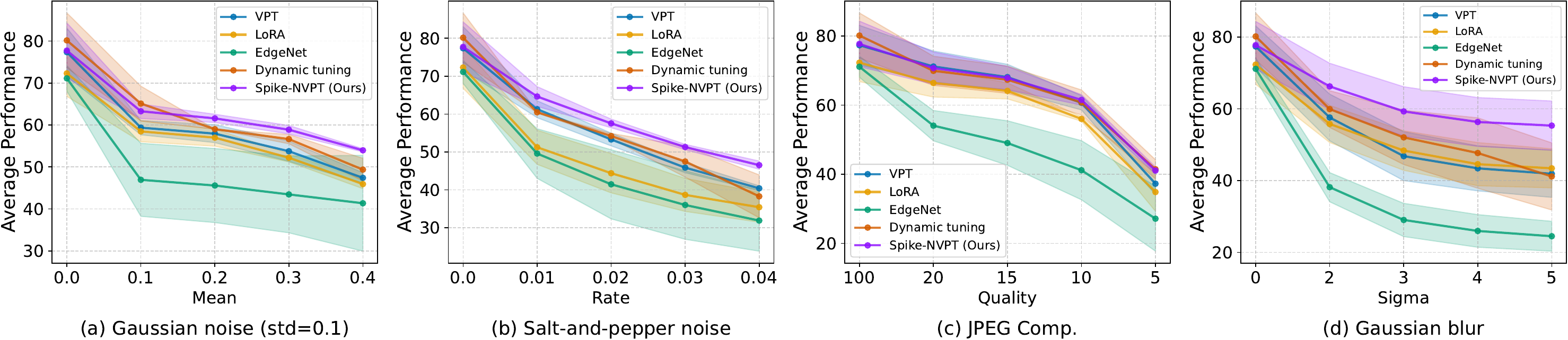}
  \caption{The average accuracies at the texture scene. The shaded area indicates the accuracy fluctuation across the two datasets in this scene.}
  \label{fig: acc on different noise level}
\end{figure*}
\begin{figure*}[htbp]
  \centering
  \includegraphics[width=1.0\linewidth]{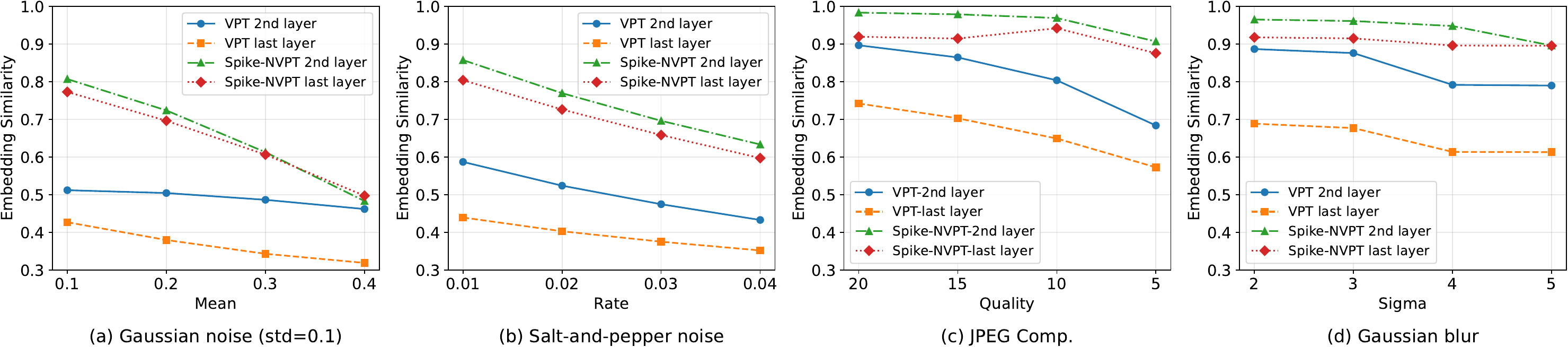}
  \caption{The similarities between the embeddings of clean and noised data at the 2nd layer and the last layer. }
  \label{fig: similarity}
\end{figure*}

\textbf{\textit{Performance with different noise degrees.}}
Figure \ref{fig: acc on different noise level} illustrates the accuracy variations of different methods when encountering varying degrees of noise in textured scenes.
The purple curve representing our method achieves the highest accuracy and the smallest fluctuation.
Suboptimal methods vary under different levels and types of noise. 
This indicates that fine-tuning approaches not specifically designed for robustness exhibit considerable differences in their resilience to various noise types.

\begin{table}[htbp]
    \centering
    \renewcommand\arraystretch{1.0}  
    \resizebox{1.0\linewidth}{!}{  
        \begin{tabular}{lccccc}
        \toprule
        \multirow{2}{*}{Methods} & \multicolumn{5}{c}{Gaussian noise (std=0.1)} \\
        \cline{2-6}
        & mean = 0.1 & mean = 0.2 & mean = 0.3 & mean = 0.4 & average \\
        \hline
        SF Layer-Only & 45.42 & 41.46 & 35.71 & 28.06 & 37.66 \\
        SD Unit-Only & \underline{48.45} & \textbf{45.76} & \underline{40.42} & \underline{33.27} & \underline{41.97} \\
        Binary VPT & 44.79 & 39.96 & 34.64 & 27.35 & 36.68 \\
        \textbf{Spike-NVPT} & \textbf{49.93} & \underline{45.59} & \textbf{41.35} & \textbf{34.54} & \textbf{43.10} \\
        \hline
        \multirow{2}{*}{Methods} & \multicolumn{5}{c}{Salt and pepper} \\
        \cline{2-6}
        & rate = 0.01 & rate = 0.02 & rate = 0.03 & rate = 0.04 & average \\
        \hline
        SF Layer-Only & 30.67 & 27.65 & 23.64 & 19.76 & 25.43 \\
        SD Unit-Only & \underline{34.67} & \underline{32.13} & \underline{29.03} & \underline{26.09} & \underline{30.48} \\
        Binary VPT & 32.42 & 29.54 & 25.51 & 22.80 & 27.57 \\
        \textbf{Spike-NVPT} & \textbf{42.73} & \textbf{36.88} & \textbf{32.45} & \textbf{29.42} & \textbf{35.37} \\
        \hline
        \multirow{2}{*}{Methods} & \multicolumn{5}{c}{JPEG compression} \\
        \cline{2-6}
        & quality = 20 & quality = 15 & quality = 10 & quality = 5 & average \\
        \hline
        SF Layer-Only & 72.90 & \underline{67.64} & 52.54 & 19.49 & 53.14 \\
        SD Unit-Only & 72.27 & 67.32 & \textbf{54.10} & 22.51 & 54.05 \\
        Binary VPT & \underline{72.94} & \textbf{68.23} & 53.68 & \underline{22.78} & \underline{54.41} \\
        \textbf{Spike-NVPT} & \textbf{73.10} & 67.51 & \underline{53.93} & \textbf{24.48} & \textbf{54.76} \\
        \hline
        \multirow{2}{*}{Methods} & \multicolumn{5}{c}{Gaussian blur} \\
        \cline{2-6}
        & sigma = 2 & sigma = 3 & sigma = 4 & sigma = 5 & average \\
        \hline
        SF Layer-Only & \underline{74.26} & 72.47 & \underline{71.41} & \underline{70.86} & \underline{72.71} \\
        SD Unit-Only & 74.00 & \textbf{73.03} & 70.83 & 70.56 & 72.10 \\
        Binary VPT & 74.24 & 71.91 & 70.78 & 70.23 & 71.79 \\
        \textbf{Spike-NVPT} & \textbf{75.51} & \underline{72.75} & \textbf{71.59} & \textbf{71.01} & \textbf{72.72} \\
        \bottomrule
        \end{tabular}
    }
    \caption{Ablation experimental results on CIFAR-100.}
    \label{tab: ablation}
\end{table}

\subsection{Ablation Study}
Ablation experiments are conducted to quantify the impacts of the Signal Filtering (SF) Layer and Spike Discretization (SD) Unit, particularly their roles in stabilizing performance and enhancing noise robustness.
Three variants are evaluated:
\begin{itemize}
    \item \textbf{SF Layer-Only}:  Uses $\mathbf{P}_{\text{Filtered}}$ (discrete values in $\{\frac{1}{T}, \frac{2}{T}, \cdots, \frac{T}{T}\}$) as the prompt.
    \item \textbf{SD Unit-Only}: Converts initialized $\mathbf{P}_{0}$ directly to spike form without signal filtering, participating in fine-tuning.
    \item \textbf{Binary VPT}: Binarizes the first layer of VPT-trained prompts using a 0.01 threshold.
\end{itemize}

Table \ref{tab: ablation} shows the ablation results.
Spike-NVPT achieves the highest average accuracy on all noise datasets.
Binary VPT exhibits degradation compared to Spike-NVPT under high-frequency noise.
It indicates naive binarization fails to enhance robustness without a dedicated fine-tuning process for binary spiking prompts.
The suboptimal performance of SD Unit-Only on high-frequency noise further demonstrates that trained binary prompts yield superior results.
While the SD Unit provides the baseline robustness, integrating the SF Layer yields the optimal performance, i.e., Spike-NVPT.
Furthermore, the performance of SF Layer-Only under Gaussian blur demonstrates that temporal integration helps preserve critical low-frequency semantic information before binarization.


\subsection{Analysis on Feature Representation Stability}
To further investigate how Spike-NVPT achieves robustness, we calculate the cosine similarity between the feature embeddings of clean and noisy images. 
A higher similarity indicates that the method successfully suppresses noise interference and maintains semantic consistency with the original input. 
Figure \ref{fig: similarity} details the similarity scores at both the second layer and the last layer under various noise conditions.

The results show that Spike-NVPT consistently yields significantly higher similarity than VPT.
Notably, when comparing similarity at shallow and deep layers, our method demonstrates exceptional performance in suppressing error propagation. 
Spike-NVPT exhibits significantly more stable trajectories, with a much smaller reduction in similarity compared to VPT.
This indicates that the spike prompts serve as stable semantic anchors, guiding the attention mechanism to persistently focus on robust features while ignoring noise-induced perturbations.

\begin{table}[htbp]
    \centering
    \renewcommand\arraystretch{1.0}  
    \resizebox{1.0\linewidth}{!}{  
    \begin{tabular}{ccccc}
        \toprule
        \multirow{1}{*}{Methods} & \makecell{Training time\\(ms/sample)} & \makecell{Inference time\\(ms/sample)} & \makecell{Memory \\usage (GB/batch)} & FLOPs(G) \\ 
        \midrule
        VPT & 1.43 & 1.38 & 7.74 & 18.51 \\
        LoRA & 4.05 & 1.32 & 8.48 & 17.71 \\
        EdgeNet & 6.31 & 5.62 & 10.65 & 24.37 \\
        Dynamic tuning & 2.99 & 0.49 & 8.90 & 17.81 \\ 
        \textbf{Spike-NVPT} & 1.40 & 1.37 & 8.37 & 18.51 \\ 
        \bottomrule
        \end{tabular}
    }
    \caption{Computational efficiency comparison of different methods on CIFAR-100.}
    \label{tab: computational efficiency}
\end{table}

\subsection{Computational efficiency of different methods} 
We compare the computational efficiency and cost of several fine-tuning methods on CIFAR-100, as shown in Table \ref{tab: computational efficiency}.
The batch sizes are uniform to 64, and the FLOPs are calculated according to the whole inference process of the model.
As the results shown in Table \ref{tab: computational efficiency}, our method does not exhibit a significant decrease in efficiency compared to traditional VPT.
This is because sparse spike prompts can reduce the computational burden of backpropagation in training, and fixed spike prompts also require no additional computation in inference.

\section{Conclusion}
This paper presents Spike-NVPT, a noise-robust visual prompt adaptation method that leverages the spiking neuron model to enhance noise resilience.
The Signal Filtering Layer in Spike-NVPT uses spiking neurons to extract critical task-specific information, converting initial continuous prompts into discrete intensity values.
A subsequent Spike Discretization Unit further transforms these discrete values into binary spike prompts.
This discrete binary information imposes a regularization-like constraint, enabling the classification head to learn more distinct class boundaries. 
Extensive experiments confirm that Spike-NVPT not only boosts the noise robustness of pre-trained visual models but also maintains competitiveness on clean samples despite the sparsity of spiking prompts.

Notably, this work marks the first application of SNN-based fine-tuning to ANN-based pre-trained models, thereby expanding the research landscape for leveraging the robustness advantages of SNNs. 
The domain transfer limitation of Spike-NVPT, as exposed in the synthetic scene, indicates future research directions. 
We will further enhance the method's performance on cross-domain tasks.


\bibliographystyle{named}
\bibliography{ijcai26}

@article{bazi2021vision,
  title={Vision transformers for remote sensing image classification},
  author={Bazi, Yakoub and Bashmal, Laila and Rahhal, Mohamad M Al and Dayil, Reham Al and Ajlan, Naif Al},
  journal={Remote Sensing},
  volume={13},
  number={3},
  pages={516},
  year={2021},
  publisher={MDPI}
}

@article{li2024learning,
  title={Learning target-aware vision transformers for real-time UAV tracking},
  author={Li, Shuiwang and Yang, Xiangyang and Wang, Xucheng and Zeng, Dan and Ye, Hengzhou and Zhao, Qijun},
  journal={IEEE Transactions on Geoscience and Remote Sensing},
  year={2024},
  publisher={IEEE}
}

@article{zhang2024segvit,
  title={Segvit v2: Exploring efficient and continual semantic segmentation with plain vision transformers},
  author={Zhang, Bowen and Liu, Liyang and Phan, Minh Hieu and Tian, Zhi and Shen, Chunhua and Liu, Yifan},
  journal={International Journal of Computer Vision},
  volume={132},
  number={4},
  pages={1126--1147},
  year={2024},
  publisher={Springer}
}

@inproceedings{
tian2024drivevlm,
title={Drive{VLM}: The Convergence of Autonomous Driving and Large Vision-Language Models},
author={Xiaoyu Tian and Junru Gu and Bailin Li and Yicheng Liu and Yang Wang and Zhiyong Zhao and Kun Zhan and Peng Jia and XianPeng Lang and Hang Zhao},
booktitle={8th Annual Conference on Robot Learning},
year={2024},
url={https://openreview.net/forum?id=928V4Umlys}
}

@article{sana2024securing,
  title={Securing the IoT Cyber Environment: Enhancing Intrusion Anomaly Detection with Vision Transformers},
  author={Sana, Laraib and Nazir, Muhammad Mohsin and Yang, Jing and Hussain, Lal and Chen, Yen-Lin and Ku, Chin Soon and Alatiyyah, Mohammed and Por, Lip Yee},
  journal={IEEE Access},
  year={2024},
  publisher={IEEE}
}

@inproceedings{guo2023robustifying,
  title={Robustifying token attention for vision transformers},
  author={Guo, Yong and Stutz, David and Schiele, Bernt},
  booktitle={Proceedings of the IEEE/CVF International Conference on Computer Vision},
  pages={17557--17568},
  year={2023}
}

@inproceedings{jia2022visual,
  title={Visual prompt tuning},
  author={Jia, Menglin and Tang, Luming and Chen, Bor-Chun and Cardie, Claire and Belongie, Serge and Hariharan, Bharath and Lim, Ser-Nam},
  booktitle={European Conference on Computer Vision},
  pages={709--727},
  year={2022},
  organization={Springer}
}

@inproceedings{
chen2023vision,
title={Vision Transformer Adapter for Dense Predictions},
author={Zhe Chen and Yuchen Duan and Wenhai Wang and Junjun He and Tong Lu and Jifeng Dai and Yu Qiao},
booktitle={The Eleventh International Conference on Learning Representations },
year={2023},
url={https://openreview.net/forum?id=plKu2GByCNW}
}

@article{maass1997networks,
  title={Networks of spiking neurons: the third generation of neural network models},
  author={Maass, Wolfgang},
  journal={Neural networks},
  volume={10},
  number={9},
  pages={1659--1671},
  year={1997},
  publisher={Elsevier}
}

@inproceedings{sharmin2020inherent,
  title={Inherent adversarial robustness of deep spiking neural networks: Effects of discrete input encoding and non-linear activations},
  author={Sharmin, Saima and Rathi, Nitin and Panda, Priyadarshini and Roy, Kaushik},
  booktitle={Computer Vision--ECCV 2020: 16th European Conference, Glasgow, UK, August 23--28, 2020, Proceedings, Part XXIX 16},
  pages={399--414},
  year={2020},
  organization={Springer}
}

@article{neftci2019surrogate,
  title={Surrogate gradient learning in spiking neural networks: Bringing the power of gradient-based optimization to spiking neural networks},
  author={Neftci, Emre O and Mostafa, Hesham and Zenke, Friedemann},
  journal={IEEE Signal Processing Magazine},
  volume={36},
  number={6},
  pages={51--63},
  year={2019},
  publisher={IEEE}
}

@inproceedings{li2020robustness,
  title={Robustness to noisy synaptic weights in spiking neural networks},
  author={Li, Chen and Chen, Runze and Moutafis, Christoforos and Furber, Steve},
  booktitle={2020 International Joint Conference on Neural Networks (IJCNN)},
  pages={1--8},
  year={2020},
  organization={IEEE}
}

@inproceedings{wang2024lion,
  title={Lion: Implicit vision prompt tuning},
  author={Wang, Haixin and Chang, Jianlong and Zhai, Yihang and Luo, Xiao and Sun, Jinan and Lin, Zhouchen and Tian, Qi},
  booktitle={Proceedings of the AAAI Conference on Artificial Intelligence},
  volume={38},
  number={6},
  pages={5372--5380},
  year={2024}
}

@book{gerstner2002spiking,
  title={Spiking neuron models: Single neurons, populations, plasticity},
  author={Gerstner, Wulfram and Kistler, Werner M},
  year={2002},
  publisher={Cambridge university press}
}

@inproceedings{
ding2024robust,
title={Robust Stable Spiking Neural Networks},
author={Jianhao Ding and Zhiyu Pan and Yujia Liu and Zhaofei Yu and Tiejun Huang},
booktitle={Forty-first International Conference on Machine Learning},
year={2024},
url={https://openreview.net/forum?id=lIYtJtpJR0}
}

@inproceedings{kundu2021hire,
  title={Hire-snn: Harnessing the inherent robustness of energy-efficient deep spiking neural networks by training with crafted input noise},
  author={Kundu, Souvik and Pedram, Massoud and Beerel, Peter A},
  booktitle={Proceedings of the IEEE/CVF International Conference on Computer Vision},
  pages={5209--5218},
  year={2021}
}

@article{ding2022snn,
  title={Snn-rat: Robustness-enhanced spiking neural network through regularized adversarial training},
  author={Ding, Jianhao and Bu, Tong and Yu, Zhaofei and Huang, Tiejun and Liu, Jian},
  journal={Advances in Neural Information Processing Systems},
  volume={35},
  pages={24780--24793},
  year={2022}
}

@article{krizhevsky2009learning,
  title={Learning multiple layers of features from tiny images},
  author={Krizhevsky, Alex and Hinton, Geoffrey and others},
  year={2009},
  publisher={Toronto, ON, Canada}
}

@inproceedings{cimpoi2014describing,
  title={Describing textures in the wild},
  author={Cimpoi, Mircea and Maji, Subhransu and Kokkinos, Iasonas and Mohamed, Sammy and Vedaldi, Andrea},
  booktitle={Proceedings of the IEEE conference on computer vision and pattern recognition},
  pages={3606--3613},
  year={2014}
}

@inproceedings{nilsback2008automated,
  title={Automated flower classification over a large number of classes},
  author={Nilsback, Maria-Elena and Zisserman, Andrew},
  booktitle={2008 Sixth Indian conference on computer vision, graphics \& image processing},
  pages={722--729},
  year={2008},
  organization={IEEE}
}

@inproceedings{parkhi2012cats,
  title={Cats and dogs},
  author={Parkhi, Omkar M and Vedaldi, Andrea and Zisserman, Andrew and Jawahar, CV},
  booktitle={2012 IEEE conference on computer vision and pattern recognition},
  pages={3498--3505},
  year={2012},
  organization={IEEE}
}

@inproceedings{zhao2024dynamic,
  title={Dynamic Tuning Towards Parameter and Inference Efficiency for ViT Adaptation},
  author={Wangbo Zhao and Jiasheng Tang and Yizeng Han and Yibing Song and Kai Wang and Gao Huang and Fan Wang and Yang You},
  booktitle={The Thirty-eighth Annual Conference on Neural Information Processing Systems},
  year={2024},
  url={https://openreview.net/forum?id=e0SQ6wsHjv}
}

@inproceedings{
hu2022lora,
title={Lo{RA}: Low-Rank Adaptation of Large Language Models},
author={Edward J Hu and yelong shen and Phillip Wallis and Zeyuan Allen-Zhu and Yuanzhi Li and Shean Wang and Lu Wang and Weizhu Chen},
booktitle={International Conference on Learning Representations},
year={2022},
url={https://openreview.net/forum?id=nZeVKeeFYf9}
}

@inproceedings{valadez2017step,
  title={The step size impact on the computational cost of spiking neuron simulation},
  author={Valadez-God{\'\i}nez, Sergio and Sossa, Humberto and Santiago-Montero, Ra{\'u}l},
  booktitle={2017 Computing Conference},
  pages={722--728},
  year={2017},
  organization={IEEE}
}

@inproceedings{schiappa2023large,
  title={A large-scale robustness analysis of video action recognition models},
  author={Schiappa, Madeline Chantry and Biyani, Naman and Kamtam, Prudvi and Vyas, Shruti and Palangi, Hamid and Vineet, Vibhav and Rawat, Yogesh S},
  booktitle={Proceedings of the IEEE/CVF conference on computer vision and pattern recognition},
  pages={14698--14708},
  year={2023}
}

@article{tsai2023convolutional,
  title={Convolutional visual prompt for robust visual perception},
  author={Tsai, Yun-Yun and Mao, Chengzhi and Yang, Junfeng},
  journal={Advances in Neural Information Processing Systems},
  volume={36},
  pages={27897--27921},
  year={2023}
}

@inproceedings{
dong2023lpt,
title={{LPT}: Long-tailed Prompt Tuning  for Image Classification},
author={Bowen Dong and Pan Zhou and Shuicheng YAN and Wangmeng Zuo},
booktitle={The Eleventh International Conference on Learning Representations },
year={2023},
url={https://openreview.net/forum?id=8pOVAeo8ie}
}

@inproceedings{zha2023instance,
  title={Instance-aware dynamic prompt tuning for pre-trained point cloud models},
  author={Zha, Yaohua and Wang, Jinpeng and Dai, Tao and Chen, Bin and Wang, Zhi and Xia, Shu-Tao},
  booktitle={Proceedings of the IEEE/CVF International Conference on Computer Vision},
  pages={14161--14170},
  year={2023}
}

@inproceedings{zhu2023visual,
  title={Visual prompt multi-modal tracking},
  author={Zhu, Jiawen and Lai, Simiao and Chen, Xin and Wang, Dong and Lu, Huchuan},
  booktitle={Proceedings of the IEEE/CVF conference on computer vision and pattern recognition},
  pages={9516--9526},
  year={2023}
}

@article{nie2023pro,
  title={Pro-tuning: Unified prompt tuning for vision tasks},
  author={Nie, Xing and Ni, Bolin and Chang, Jianlong and Meng, Gaofeng and Huo, Chunlei and Xiang, Shiming and Tian, Qi},
  journal={IEEE Transactions on Circuits and Systems for Video Technology},
  volume={34},
  number={6},
  pages={4653--4667},
  year={2023},
  publisher={IEEE}
}

@inproceedings{zheng2021visual,
  title={Visual language based succinct zero-shot object detection},
  author={Zheng, Ye and Huang, Xi and Cui, Li},
  booktitle={Proceedings of the 29th ACM International Conference on Multimedia},
  pages={5410--5418},
  year={2021}
}

@inproceedings{zhou2024bsbp,
  title={Bsbp-rwkv: Background suppression with boundary preservation for efficient medical image segmentation},
  author={Zhou, Xudong and Chen, Tianxiang},
  booktitle={Proceedings of the 32nd ACM International Conference on Multimedia},
  pages={4938--4946},
  year={2024}
}

@inproceedings{liu2023icmh,
  title={Icmh-net: Neural image compression towards both machine vision and human vision},
  author={Liu, Lei and Hu, Zhihao and Chen, Zhenghao and Xu, Dong},
  booktitle={Proceedings of the 31st ACM International Conference on Multimedia},
  pages={8047--8056},
  year={2023}
}

@article{zhan2024two,
  title={A two-stage spiking meta-learning method for few-shot classification},
  author={Zhan, Qiugang and Wang, Bingchao and Jiang, Anning and Xie, Xiurui and Zhang, Malu and Liu, Guisong},
  journal={Knowledge-Based Systems},
  volume={284},
  pages={111220},
  year={2024},
  publisher={Elsevier}
}

@article{zhan2024spiking,
  title={Spiking transfer learning from rgb image to neuromorphic event stream},
  author={Zhan, Qiugang and Liu, Guisong and Xie, Xiurui and Tao, Ran and Zhang, Malu and Tang, Huajin},
  journal={IEEE Transactions on Image Processing},
  year={2024},
  publisher={IEEE}
}

@article{liao2023convolutional,
  title={A convolutional spiking neural network with adaptive coding for motor imagery classification},
  author={Liao, Xiaojian and Wu, Yuli and Wang, Zi and Wang, Deheng and Zhang, Hongmiao},
  journal={Neurocomputing},
  volume={549},
  pages={126470},
  year={2023},
  publisher={Elsevier}
}

@article{ding2025neuromorphic,
  title={Neuromorphic computing paradigms enhance robustness through spiking neural networks},
  author={Ding, Jianhao and Yu, Zhaofei and Liu, Jian K and Huang, Tiejun},
  journal={Nature Communications},
  volume={16},
  number={1},
  pages={10175},
  year={2025},
  publisher={Nature Publishing Group UK London}
}

@inproceedings{wang2023new,
  title={A new ann-snn conversion method with high accuracy, low latency and good robustness},
  author={Wang, Bingsen and Cao, Jian and Chen, Jue and Feng, Shuo and Wang, Yuan},
  booktitle={Proceedings of the Thirty-Second International Joint Conference on Artificial Intelligence},
  pages={3067--3075},
  year={2023}
}

@article{fei2007learning,
  title={Learning generative visual models from few training examples: An incremental Bayesian approach tested on 101 object categories},
  author={Fei-Fei, Li and Fergus, Rob and Perona, Pietro},
  journal={Computer vision and Image understanding},
  volume={106},
  number={1},
  pages={59--70},
  year={2007},
  publisher={Elsevier}
}

@article{zhai2019large,
  title={A large-scale study of representation learning with the visual task adaptation benchmark},
  author={Zhai, Xiaohua and Puigcerver, Joan and Kolesnikov, Alexander and Ruyssen, Pierre and Riquelme, Carlos and Lucic, Mario and Djolonga, Josip and Pinto, Andre Susano and Neumann, Maxim and Dosovitskiy, Alexey and others},
  journal={arXiv preprint arXiv:1910.04867},
  year={2019}
}

@article{Cheng_2017,
   title={Remote Sensing Image Scene Classification: Benchmark and State of the Art},
   volume={105},
   ISSN={1558-2256},
   url={http://dx.doi.org/10.1109/JPROC.2017.2675998},
   DOI={10.1109/jproc.2017.2675998},
   number={10},
   journal={Proceedings of the IEEE},
   publisher={Institute of Electrical and Electronics Engineers (IEEE)},
   author={Cheng, Gong and Han, Junwei and Lu, Xiaoqiang},
   year={2017},
   month={Oct},
   pages={1865-1883}
}

@inproceedings{li2024harnessing,
  title={Harnessing edge information for improved robustness in vision transformers},
  author={Li, Yanxi and Du, Chengbin and Xu, Chang},
  booktitle={Proceedings of the AAAI Conference on Artificial Intelligence},
  volume={38},
  number={4},
  pages={3252--3260},
  year={2024}
}

@article{doi:10.1126/sciadv.adi1480,
    author = {Wei Fang  and Yanqi Chen  and Jianhao Ding  and Zhaofei Yu  and Timothée Masquelier  and Ding Chen  and Liwei Huang  and Huihui Zhou  and Guoqi Li  and Yonghong Tian },
    title = {SpikingJelly: An open-source machine learning infrastructure platform for spike-based intelligence},
    journal = {Science Advances},
    volume = {9},
    number = {40},
    pages = {eadi1480},
    year = {2023},
    doi = {10.1126/sciadv.adi1480},
    URL = {https://www.science.org/doi/abs/10.1126/sciadv.adi1480},
    eprint = {https://www.science.org/doi/pdf/10.1126/sciadv.adi1480},
}
\onecolumn  
{
\noindent{\large\bfseries Supplementary Material}\\[5pt]
Table \ref{tab: CIFAR-100 noise results}--\ref{tab: DMLab noise results}
record the complete test results of Spike-NVPT and other baseline methods across seven datasets.

\centering
\renewcommand\arraystretch{1.0}
\resizebox{\linewidth}{!}{%
\begin{tabular}{lccccc|ccccc}
\toprule
\multirow{2}{*}{Methods} & \multicolumn{5}{c|}{Gaussian noise(std=0.1)} & \multicolumn{5}{c}{JPEG compression} \\
\cline{2-11}
& mean=0.1 & mean=0.2 & mean=0.3 & mean=0.4 & average & quality=20 & quality=15 & quality=10 & quality=5 & average \\
\hline
VPT & 43.24 & 39.36 & 34.01 & 26.81 & 35.86 & \underline{69.23} & 63.59 & 48.87 & 15.86 & 49.39 \\
LoRA & 42.57 & 40.00 & 36.05 & 29.90 & 37.13 & 66.33 & 61.44 & 47.63 & 20.77 & 49.04 \\
EdgeNet & \textbf{50.58} & \textbf{48.09} & \textbf{44.84} & \textbf{38.73} & \textbf{45.56} & 55.22 & 49.16 & 39.74 & \underline{23.01} & 41.78 \\
Dynamic Tuning & 43.15 & 41.05 & 28.54 & 26.90 & 34.91 & 68.61 & \underline{66.55} & \underline{52.23} & 17.64 & \underline{51.26} \\
\textbf{Spike-NVPT (Ours)} & \underline{49.93} & \underline{46.59} & \underline{41.35} & \underline{34.54} & \underline{43.10} & \textbf{73.10} & \textbf{67.51} & \textbf{53.93} & \textbf{24.48} & \textbf{54.76} \\
\hline
\multirow{2}{*}{Methods} & \multicolumn{5}{c|}{Salt and pepper} & \multicolumn{5}{c}{Gaussian blur} \\
\cline{2-11}
& rate=0.01 & rate=0.02 & rate=0.03 & rate=0.04 & average & sigma=2 & sigma=3 & sigma=4 & sigma=5 & average \\
\hline
VPT & 30.54 & 25.89 & 20.29 & 16.93 & 23.41 & 68.62 & 65.85 & 65.56 & \underline{64.13} & 66.04 \\
LoRA & 30.47 & 30.68 & 29.30 & \underline{27.86} & 29.58 & 67.44 & 65.84 & 64.69 & 63.98 & 65.49 \\
EdgeNet & \underline{48.37} & 35.89 & 28.71 & 24.78 & 34.44 & 65.20 & 63.13 & 62.72 & 62.12 & 63.29 \\
Dynamic Tuning & \textbf{53.65} & \textbf{42.41} & \textbf{34.53} & 26.21 & \textbf{39.20} & \underline{70.61} & \underline{67.19} & \underline{66.63} & 61.18 & \underline{66.40} \\
\textbf{Spike-NVPT (Ours)} & 42.73 & \underline{36.88} & \underline{32.45} & \textbf{29.42} & \underline{35.37} & \textbf{75.51} & \textbf{72.75} & \textbf{71.59} & \textbf{71.01} & \textbf{72.72} \\
\bottomrule
\end{tabular}%
}
\captionof{table}{Performance comparison of different methods on the CIFAR-100 dataset with varying noise.}
\label{tab: CIFAR-100 noise results}

\centering
\renewcommand\arraystretch{1.0}
\resizebox{\linewidth}{!}{%
        \begin{tabular}{lccccc|ccccc}
        \toprule
        \multirow{2}{*}{Methods} & \multicolumn{5}{c|}{Gaussian noise(std=0.1)} & \multicolumn{5}{c}{JPEG compression} \\
        \cline{2-11}
        & mean=0.1 & mean=0.2 & mean=0.3 & mean=0.4 & average & quality=20 & quality=15 & quality=10 & quality=5 & average \\
        \hline
        VPT & 81.41 & 79.08 & 74.18 & 64.73 & 74.85 & 86.75 & 86.08 & 83.83 & 67.61 & 81.07 \\
        LoRA & 83.42 & 82.58 & 81.41 & 78.26 & 81.42 & 86.47 & 85.95 & 85.03 & \underline{76.17} & 83.41 \\
        EdgeNet & 81.35 & 81.30 & 80.67 & 79.77 & 80.77 & 81.43 & 80.85 & 78.16 & 68.36 & 77.20 \\
        Dynamic Tuning & \textbf{87.39} & \textbf{86.90} & \underline{83.86} & \underline{80.02} & \underline{84.54} & \textbf{87.44} & \underline{86.62} & \textbf{86.39} & \textbf{80.03} & \textbf{85.12} \\
        \textbf{Spike-NVPT (Ours)} & \underline{87.08} & \underline{86.15} & \textbf{84.40} & \textbf{81.35} & \textbf{84.75} & \underline{87.38} & \textbf{86.66} & \underline{85.14} & 75.02 & \underline{83.55} \\
        \hline
        \multirow{2}{*}{Methods} & \multicolumn{5}{c|}{Salt and pepper} & \multicolumn{5}{c}{Gaussian blur} \\
        \cline{2-11}
         & rate=0.01 & rate=0.02 & rate=0.03 & rate=0.04 & average & sigma=2 & sigma=3 & sigma=4 & sigma=5 & average \\
        \hline
        VPT & 84.91 & 79.52 & 74.86 & 69.86 & 77.29 & 83.24 & 71.91 & 65.70 & 64.11 & 71.24 \\
        LoRA & 83.42 & 80.03 & 77.29 & 73.79 & 78.63 & \textbf{84.95} & \textbf{79.82} & \textbf{77.88} & \textbf{77.49} & \textbf{80.04} \\
        EdgeNet & 81.53 & 75.32 & 69.81 & 65.51 & 73.04 & 76.47 & 69.83 & 67.03 & 66.13 & 69.87 \\
        Dynamic Tuning & \textbf{87.48} & \textbf{83.12} & \underline{78.52} & \underline{76.31} & \underline{81.36} & 81.71 & 77.29 & \underline{74.77} & 70.98 & 76.19 \\
        \textbf{Spike-NVPT (Ours)} & \underline{86.33} & \underline{83.09} & \textbf{79.64} & \textbf{76.43} & \textbf{81.37} & \underline{84.63} & \underline{77.62} & 74.05 & \underline{72.85} & \underline{77.29} \\
        \bottomrule
        \end{tabular}
    }
    \captionof{table}{Performance comparison of different methods on the Caltech-101 dataset with varying noise. }
    \label{tab: Caltech-101 noise results}

\centering
\renewcommand\arraystretch{1.0}
\resizebox{\linewidth}{!}{%
        \begin{tabular}{lccccc|ccccc}
        \toprule
        \multirow{2}{*}{Methods} & \multicolumn{5}{c|}{Gaussian noise(std=0.1)} & \multicolumn{5}{c}{JPEG compression} \\
        \cline{2-11}
        & mean=0.1 & mean=0.2 & mean=0.3 & mean=0.4 & average & quality=20 & quality=15 & quality=10 & quality=5 & average \\
        \hline
        VPT & 77.57 & 76.51 & 72.31 & 67.70 & 73.52 & 82.28 & 79.72 & 73.21 & 48.43 & 70.91 \\
        LoRA & 80.78 & 79.15 & 76.37 & 71.46 & 76.94 & 79.78 & 78.55 & 73.26 & 53.69 & 71.32 \\
        EdgeNet & 81.82 & 82.34 & 80.81 & \textbf{79.12} & 81.02 & 78.39 & 76.72 & 73.51 & \textbf{61.54} & 72.54 \\
        Dynamic Tuning & \textbf{84.08} & \underline{83.44} & \textbf{81.60} & \underline{78.23} & \textbf{81.84} & \underline{83.56} & \underline{81.55} & 75.52 & \underline{59.06} & \underline{74.92} \\
        \textbf{Spike-NVPT (Ours)} & \underline{84.00} & \textbf{83.46} & \underline{80.92} & 77.02 & \underline{81.35} & \textbf{83.89} & \textbf{81.96} & \textbf{77.60} & 56.72 & \textbf{75.04} \\
        \hline
        \multirow{2}{*}{Methods} & \multicolumn{5}{c|}{Salt and pepper} & \multicolumn{5}{c}{Gaussian blur} \\
        \cline{2-11}
         & rate=0.01 & rate=0.02 & rate=0.03 & rate=0.04 & average & sigma=2 & sigma=3 & sigma=4 & sigma=5 & average \\
        \hline
        VPT & 77.54 & 69.69 & 63.04 & 58.52 & 67.20 & 76.81 & 68.27 & 64.68 & 63.10 & 68.22 \\
        LoRA & 80.87 & 75.28 & 68.06 & 62.61 & 71.71 & 78.52 & \underline{72.44} & \underline{69.20} & \underline{68.47} & \underline{72.16} \\
        EdgeNet & 82.12 & 76.34 & 71.11 & 67.59 & 74.29 & 79.45 & 71.44 & 67.02 & 65.41 & 70.83 \\
        Dynamic Tuning & \textbf{84.68} & \underline{77.60} & \textbf{73.94} & \underline{68.25} & \underline{76.12} & \underline{80.92} & 70.78 & 62.82 & 54.02 & 67.14 \\
        \textbf{Spike-NVPT (Ours)} & \underline{83.57} & \textbf{78.20} & \underline{73.32} & \textbf{69.80} & \textbf{76.22} & \textbf{82.26} & \textbf{77.95} & \textbf{75.39} & \textbf{74.11} & \textbf{77.43} \\
        \bottomrule
        \end{tabular}
    }
     \captionof{table}{Performance comparison of different methods on the Pets dataset with varying noise. }
    \label{tab: Pets noise results}

\centering
\renewcommand\arraystretch{1.0}
\resizebox{\linewidth}{!}{%
        \begin{tabular}{lccccc|ccccc}
        \toprule
        \multirow{2}{*}{Methods} & \multicolumn{5}{c|}{Gaussian noise(std=0.1)} & \multicolumn{5}{c}{JPEG compression} \\
        \cline{2-11}
        & mean=0.1 & mean=0.2 & mean=0.3 & mean=0.4 & average & quality=20 & quality=15 & quality=10 & quality=5 & average \\
        \hline
        VPT & \underline{95.56} & \textbf{95.17} & \textbf{93.01} & \textbf{89.30} & \textbf{93.26} & \textbf{97.64} & \textbf{96.85} & \underline{93.35} & 69.93 & 89.44 \\
        LoRA & 94.88 & 93.67 & 91.61 & 86.81 & 91.74 & \underline{97.38} & \underline{96.54} & \textbf{93.87} & \textbf{76.06} & \textbf{90.96} \\
        EdgeNet & 90.84 & 90.99 & 90.18 & \underline{88.47} & 90.12 & 92.60 & 89.45 & 81.12 & 56.55 & 79.93 \\
        Dynamic Tuning & \textbf{95.62} & \underline{94.54} & 88.84 & 88.29 & 91.82 & 93.92 & 92.06 & 89.28 & 72.27 & 86.88 \\
        \textbf{Spike-NVPT (Ours)} & 94.76 & 93.90 & \underline{91.97} & 87.43 & \underline{92.02} & 97.37 & 96.68 & 93.66 & \underline{74.96} & \underline{90.67} \\
        \hline
        \multirow{2}{*}{Methods} & \multicolumn{5}{c|}{Salt and pepper} & \multicolumn{5}{c}{Gaussian blur} \\
        \cline{2-11}
         & rate=0.01 & rate=0.02 & rate=0.03 & rate=0.04 & average & sigma=2 & sigma=3 & sigma=4 & sigma=5 & average \\
        \hline
        VPT & \underline{95.74} & \textbf{92.19} & \underline{86.78} & \underline{81.98} & \textbf{89.17} & \underline{96.47} & \textbf{94.44} & \textbf{93.10} & \textbf{92.44} & \textbf{94.11} \\
        LoRA & 94.75 & 90.00 & 83.43 & 78.22 & 86.60 & 95.93 & 92.50 & 90.91 & 90.45 & 92.45 \\
        EdgeNet & 89.92 & 84.09 & 78.87 & 73.90 & 81.70 & 87.98 & 78.96 & 75.00 & 73.38 & 78.83 \\
        Dynamic Tuning & \textbf{96.44} & \underline{91.76} & 85.71 & 76.92 & 87.71 & 88.18 & 83.61 & 79.35 & 72.06 & 80.80 \\
        \textbf{Spike-NVPT (Ours)} & 95.10 & 91.62 & \textbf{87.07} & \textbf{82.84} & \underline{89.16} & \textbf{96.52} & \underline{93.95} & \underline{92.73} & \underline{91.97} & \underline{93.79} \\
        \bottomrule
        \end{tabular}
    }
    \captionof{table}{Performance comparison of different methods on the Flowers dataset with varying noise. }
    \label{tab: Flowers noise results}

\centering
\renewcommand\arraystretch{1.0}
\resizebox{\linewidth}{!}{%
        \begin{tabular}{lccccc|ccccc}
        \toprule
        \multirow{2}{*}{Methods} & \multicolumn{5}{c|}{Gaussian noise(std=0.1)} & \multicolumn{5}{c}{JPEG compression} \\
        \cline{2-11}
        & mean=0.1 & mean=0.2 & mean=0.3 & mean=0.4 & average & quality=20 & quality=15 & quality=10 & quality=5 & average \\
        \hline
        VPT & 57.66 & 55.80 & 51.33 & 46.17 & 52.74 & \textbf{66.70} & \underline{64.41} & \underline{58.62} & 40.43 & 57.54 \\
        LoRA & 56.01 & 54.95 & 51.22 & 47.02 & 52.30 & 62.34 & 61.81 & 55.37 & 40.43 & 54.99 \\
        EdgeNet & 55.64 & 54.41 & 52.61 & \underline{52.77} & 53.86 & 58.46 & 55.48 & 49.73 & 36.54 & 50.05 \\
        Dynamic Tuning & \underline{60.96} & \underline{58.99} & \underline{55.53} & 52.07 & \underline{56.89} & 65.70 & 63.42 & 57.19 & \textbf{44.21} & \underline{57.63} \\
        \textbf{Spike-NVPT (Ours)} & \textbf{61.54} & \textbf{60.59} & \textbf{57.87} & \textbf{54.36} & \textbf{58.59} & \underline{66.06} & \textbf{64.52} & \textbf{60.00} & \underline{42.13} & \textbf{58.18} \\
        \hline
        \multirow{2}{*}{Methods} & \multicolumn{5}{c|}{Salt and pepper} & \multicolumn{5}{c}{Gaussian blur} \\
        \cline{2-11}
         & rate=0.01 & rate=0.02 & rate=0.03 & rate=0.04 & average & sigma=2 & sigma=3 & sigma=4 & sigma=5 & average \\
        \hline
        VPT & 59.04 & 51.60 & 44.52 & 39.73 & 48.72 & 51.06 & 40.00 & 37.13 & 35.32 & 40.88 \\
        LoRA & 55.74 & 49.57 & 43.03 & 39.36 & 46.93 & 50.59 & 42.87 & \underline{38.56} & \underline{37.98} & \underline{42.50} \\
        EdgeNet & 56.12 & 50.59 & 45.05 & 39.95 & 47.93 & 42.23 & 33.67 & 30.53 & 28.67 & 33.78 \\
        Dynamic Tuning & \underline{60.64} & \underline{53.51} & \textbf{51.60} & \underline{43.94} & \underline{52.42} & \underline{54.79} & \underline{44.47} & 37.87 & 31.81 & 42.23 \\
        \textbf{Spike-NVPT (Ours)} & \textbf{62.02} & \textbf{56.38} & \underline{50.59} & \textbf{45.43} & \textbf{53.61} & \textbf{59.79} & \textbf{52.39} & \textbf{49.52} & \textbf{48.46} & \textbf{52.54} \\
        \bottomrule
        \end{tabular}
    }
    \captionof{table}{Performance comparison of different methods on the DTD dataset with varying noise. }
    \label{tab: DTD noise results}

\centering
\renewcommand\arraystretch{1.0}
\resizebox{\linewidth}{!}{%
        \begin{tabular}{lccccc|ccccc}
        \toprule
        \multirow{2}{*}{Methods} & \multicolumn{5}{c|}{Gaussian noise(std=0.1)} & \multicolumn{5}{c}{JPEG compression} \\
        \cline{2-11}
        & mean=0.1 & mean=0.2 & mean=0.3 & mean=0.4 & average & quality=20 & quality=15 & quality=10 & quality=5 & average \\
        \hline
        VPT & 61.08 & \underline{60.08} & 56.17 & \underline{48.65} & 56.50 & \textbf{75.68} & \textbf{71.83} & 62.78 & 34.10 & 61.10 \\
        LoRA & 60.95 & 59.02 & 53.19 & 44.84 & 54.50 & 70.51 & 66.46 & 56.67 & 29.29 & 55.73 \\
        EdgeNet & 38.25 & 36.76 & 34.32 & 29.95 & 34.82 & 49.68 & 42.60 & 32.62 & 17.75 & 35.66 \\
        Dynamic Tuning & \textbf{69.25} & 59.03 & \underline{57.75} & 46.67 & \underline{58.17} & 74.16 & \underline{71.30} & \textbf{64.41} & \underline{38.70} & \underline{62.14} \\
        \textbf{Spike-NVPT (Ours)} & \underline{65.00} & \textbf{62.56} & \textbf{59.89} & \textbf{53.59} & \textbf{60.26} & \underline{75.38} & 71.22 & \underline{63.05} & \textbf{40.02} & \textbf{62.42} \\
        \hline
        \multirow{2}{*}{Methods} & \multicolumn{5}{c|}{Salt and pepper} & \multicolumn{5}{c}{Gaussian blur} \\
        \cline{2-11}
         & rate=0.01 & rate=0.02 & rate=0.03 & rate=0.04 & average & sigma=2 & sigma=3 & sigma=4 & sigma=5 & average \\
        \hline
        VPT & \underline{63.51} & \underline{54.98} & \underline{47.19} & \underline{41.06} & \underline{51.69} & 64.08 & 53.62 & 49.70 & 48.48 & 53.97 \\
        LoRA & 46.73 & 39.19 & 34.30 & 31.46 & 37.92 & 60.95 & 53.86 & 50.59 & 48.86 & 53.57 \\
        EdgeNet & 43.03 & 32.33 & 26.94 & 23.81 & 31.53 & 34.10 & 24.41 & 21.43 & 20.35 & 25.07 \\
        Dynamic Tuning & 60.30 & 54.89 & 43.35 & 32.62 & 47.79 & \underline{65.03} & \underline{59.60} & \underline{57.49} & \underline{50.51} & \underline{58.16} \\
        \textbf{Spike-NVPT (Ours)} & \textbf{67.24} & \textbf{58.68} & \textbf{52.00} & \textbf{47.60} & \textbf{56.38} & \textbf{72.68} & \textbf{66.14} & \textbf{63.13} & \textbf{62.17} & \textbf{66.03} \\
        \bottomrule
        \end{tabular}
    }
    \captionof{table}{Performance comparison of different methods on the RESISC45 dataset with varying noise. }
    \label{tab: RESISC45 noise results}

\centering
\renewcommand\arraystretch{1.0}
\resizebox{\linewidth}{!}{%
        \begin{tabular}{lccccc|ccccc}
        \toprule
        \multirow{2}{*}{Methods} & \multicolumn{5}{c|}{Gaussian noise(std=0.1)} & \multicolumn{5}{c}{JPEG compression} \\
        \cline{2-11}
        & mean=0.1 & mean=0.2 & mean=0.3 & mean=0.4 & average & quality=20 & quality=15 & quality=10 & quality=5 & average \\
        \hline
        VPT & 40.27 & 40.29 & 39.59 & 38.79 & 39.74 & 41.10 & 40.65 & 39.40 & \underline{35.06} & 39.05 \\
        LoRA & 40.44 & 40.40 & 39.07 & 37.59 & 39.38 & 44.06 & \textbf{42.16} & 38.83 & 33.62 & 39.67 \\
        EdgeNet & 32.65 & 32.03 & 30.70 & 29.56 & 31.24 & 35.38 & 34.91 & 32.37 & 28.75 & 32.85 \\
        Dynamic Tuning & \textbf{45.31} & \textbf{44.34} & \textbf{42.98} & \textbf{40.43} & \textbf{43.27} & \textbf{46.13} & 41.28 & \textbf{40.13} & \textbf{36.66} & \textbf{41.08} \\
        \textbf{Spike-NVPT (Ours)} & \underline{40.71} & \underline{41.17} & \underline{41.21} & \underline{40.37} & \underline{40.87} & \underline{44.46} & \underline{42.00} & \underline{39.98} & 33.78 & \underline{40.06} \\
        \hline
        \multirow{2}{*}{Methods} & \multicolumn{5}{c|}{Salt and pepper} & \multicolumn{5}{c}{Gaussian blur} \\
        \cline{2-11}
         & rate=0.01 & rate=0.02 & rate=0.03 & rate=0.04 & average & sigma=2 & sigma=3 & sigma=4 & sigma=5 & average \\
        \hline
        VPT & 40.22 & 39.44 & 38.54 & 38.31 & 39.13 & 36.16 & 32.97 & 31.64 & 31.12 & 32.97 \\
        LoRA & \underline{43.67} & 40.91 & 38.80 & 37.29 & 40.17 & \textbf{41.15} & \underline{38.55} & \textbf{36.78} & \textbf{36.18} & \textbf{38.17} \\
        EdgeNet & 30.87 & 28.92 & 26.69 & 25.67 & 28.04 & 29.82 & 27.25 & 26.47 & 25.97 & 27.38 \\
        Dynamic Tuning & \textbf{45.84} & \textbf{44.97} & \textbf{42.79} & \underline{38.92} & \textbf{43.14} & \underline{40.95} & \textbf{38.89} & \underline{35.46} & \underline{34.14} & \underline{37.36} \\
        \textbf{Spike-NVPT (Ours)} & 42.23 & \underline{41.62} & \underline{40.65} & \textbf{39.23} & \underline{40.93} & 38.70 & 35.32 & 33.68 & 32.96 & 35.17 \\
        \bottomrule
        \end{tabular}
    }
    \captionof{table}{Performance comparison of different methods on the DMLab dataset with varying noise. }
    \label{tab: DMLab noise results}
\vspace{1em}
}
\twocolumn  


\end{document}